\documentclass[letterpaper, 10 pt, conference]{ieeeconf}  

\IEEEoverridecommandlockouts                              

\usepackage{times}
\usepackage{epsfig}
\usepackage{graphicx}
\usepackage{amsmath}
\usepackage{amssymb}
\usepackage[utf8]{inputenc}
\usepackage{textgreek}
\usepackage[table]{xcolor}
\usepackage{arydshln}

\usepackage{amsmath}
\usepackage{amssymb}
\usepackage{algorithm}
\usepackage{algpseudocode}

\usepackage{booktabs}
\usepackage{multirow}
\usepackage{bm}
\usepackage{threeparttable}
\usepackage{algorithm}
\usepackage{algpseudocode}
\usepackage{cite}
\usepackage[breaklinks=true,bookmarks=false]{hyperref}
\hypersetup
{
colorlinks = true,
linkcolor = red,
anchorcolor = black,
citecolor = green,
urlcolor = magenta
}

\hypersetup{
  pdftitle={Your Paper Title},
  pdfauthor={},
  pdfsubject={},
  pdfkeywords={}
}

\title{
TAO-Force: Unifying Force-Aware Perception and Fast--Slow Control
for Contact-Rich Manipulation
}

\author{
Bohan Gan, Xuanzhang Wen, Yongsheng Zhao,
Baoping Cheng\textsuperscript{*},\\
Wenhe Jia, Ye Wang, Gongxin Yao, Han Gao,
Jingyao Tang, Lei Zhao, and Ji Ge\\[0.4em]
China Mobile (Hangzhou) Information Technology Co., Ltd.,
Hangzhou 310023, China\\[0.25em]
{\small\ttfamily
\{ganbohan, wenxuanzhang, zhaoyongsheng, chengbaoping, jiawenhe,}\\
{\small\ttfamily
wangye, yaogongxin, gaohan, tangjingyao, zhaolei,
geji\}@cmhi.chinamobile.com}\\[0.25em]
{\small
\textsuperscript{*}Corresponding author: Baoping Cheng}
}

\begin{document}

\maketitle
\thispagestyle{empty}
\pagestyle{empty}

\begin {abstract}
Vision-Language-Action (VLA) models have demonstrated strong performance across diverse robotic manipulation tasks, yet their predominantly vision-centric perception and position-controlled execution remain insufficient for contact-rich manipulation. Visual observations alone often provide limited evidence of contact onset and interaction magnitude, while position-control policies cannot respond compliantly to rapidly changing contact dynamics. To bridge both the perception and control gaps, we propose \textbf {TAO-Force}, a force-conditioned VLA framework that combines force-aware policy learning with contact-regulated execution. For force-aware perception, TAO-Force introduces Force-conditioned Feature-wise Linear Modulation (F-FiLM) to inject encoded force feedback into the representations of a frozen pretrained visual-language backbone while preserving its semantic priors. For responsive control, it employs a contact-gated fast--slow architecture, with a slow position-control branch tracking nominal trajectories during non-contact phases and a fast admittance-control branch regulating physical interaction during contact phases. Detailed analyses on a force-perception task and real-world evaluations across four contact-rich manipulation tasks validate the effectiveness and robustness of TAO-Force.
\end {abstract}

\section{Introduction}

Vision-Language-Action (VLA) models have emerged as a powerful paradigm for embodied manipulation, unifying perception, planning, and action generation within an end-to-end architecture~\cite{brohan2022rt,zitkovich2023rt,kim2024openvla,black2024pi_0,liu2024rdt}. By leveraging the semantic priors and generalization capabilities from large-scale pretraining, VLA models have demonstrated strong performance across diverse open-world tasks in both simulation and real-world settings~\cite{intelligence2025pi05,bjorck2025gr00t}. However, their predominantly vision-centric design faces a fundamental limitation in contact-rich manipulation. In tasks such as peeling, pressing, and surface wiping, critical state transitions are governed by physical interaction rather than visual appearance. Contact onset, interaction magnitude, and the compliance required for success may correspond to only subtle visual changes. Consequently, policies that rely primarily on visual observations can be insensitive to the physical-interaction signals that determine task feasibility and execution robustness. Bridging this perception gap requires treating force sensing\footnote{Throughout this paper, unless otherwise specified, the term \emph{force} denotes the six-dimensional end-effector wrench comprising three-dimensional force and three-dimensional torque.} as a first-class modality, allowing the policy to reason about interaction states that are difficult to infer from visual observations alone~\cite{zhang2025elucidating,yu2026forcevla,2024FoAR,lee2026manipforce}.
\begin{figure}[t]
    \centering
    \includegraphics[width=\linewidth]{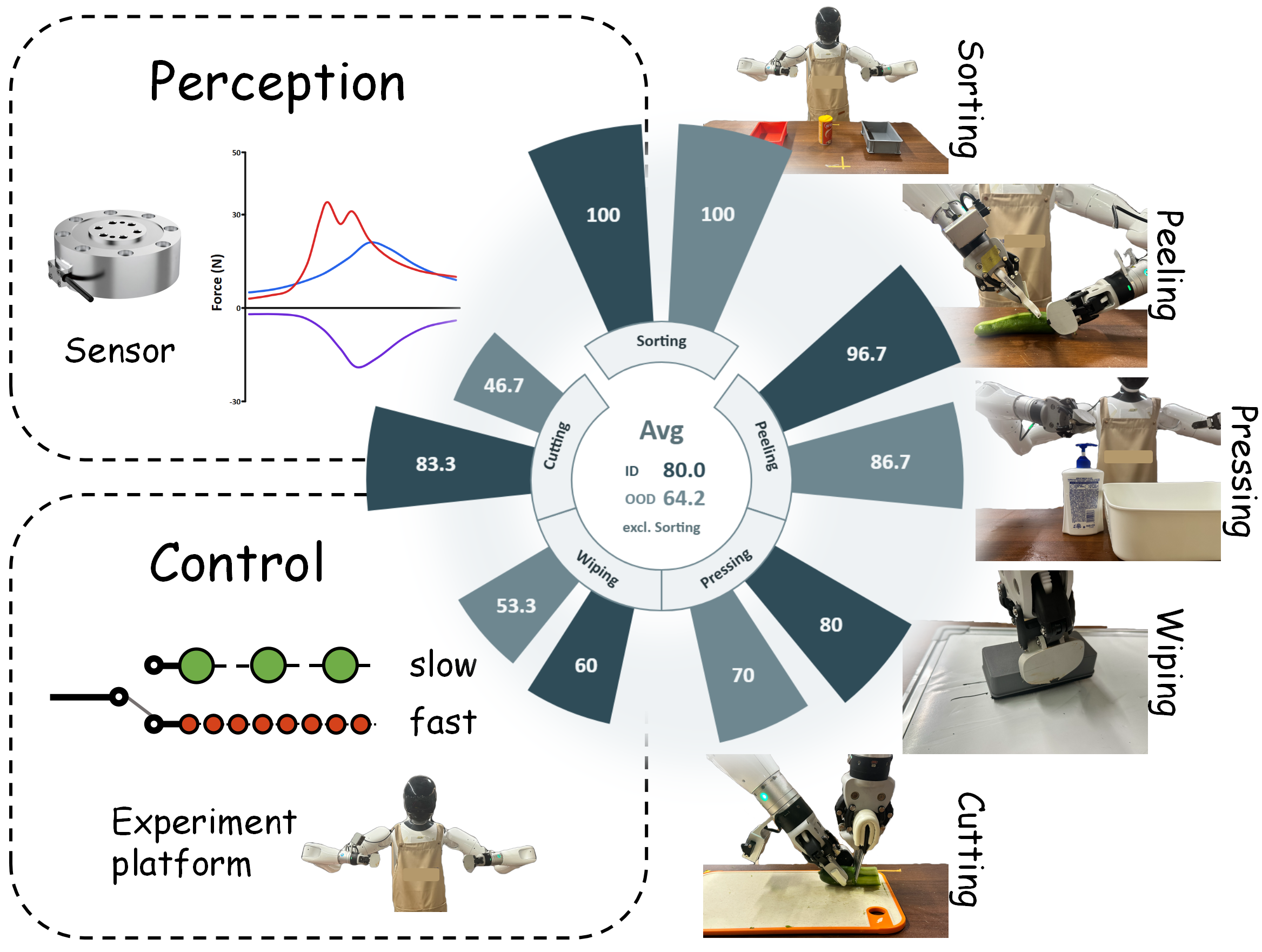} 
    \caption{\textbf{System overview and real-world evaluation.} Wrist-mounted six-axis force/torque sensors provide force feedback for policy perception. At execution time, a fast force-feedback loop updates joint commands in contact phases at a higher frequency than the slow position-control loop in non-contact phases. We evaluate TAO-Force on one force-perception task and four contact-rich manipulation tasks, achieving average success rates of 80.0\% and 64.2\% on the contact-rich tasks under in-domain (ID) and out-of-domain (OOD) conditions, respectively.}
    \label{fig:first_img}
\end{figure}

Recent studies have begun incorporating force feedback into VLA frameworks. Early approaches condition action generation on joint-torque signals or end-effector forces while retaining position commands as the policy output~\cite{zhang2025elucidating,yu2026forcevla}. However, when force is used only as an input and the predicted position commands are directly executed, the system may not respond sufficiently quickly to rapidly changing contact dynamics, limiting contact compliance and execution robustness. Subsequent methods therefore jointly predict motion and force targets, which are tracked by higher-frequency low-level controllers during contact~\cite{li2026forcevla2,liu2025forcemimic}. Despite their promising performance, two key challenges remain. First, unlike vision, force sensing lacks a feature extractor pretrained at a comparable scale. During post-training, pretrained visual representations can dominate optimization, causing the policy to underutilize force information that is essential for contact regulation. Second, existing joint motion--force prediction strategies generally provide control targets without explicitly estimating contact states or coordinating transitions between non-contact and contact phases, making it difficult to balance motion accuracy with contact compliance.

To address these challenges, we propose TAO-Force, a force-conditioned adaptation framework that bridges the perception and control gaps in contact-rich VLA deployment. As illustrated in Fig.~\ref{fig:first_img}, TAO-Force retains a frozen pretrained VLA backbone and incorporates physical-interaction feedback into policy learning and execution through two complementary designs. First, we introduce \textbf{F}orce-conditioned \textbf{F}eature-w\textbf{i}se \textbf{L}inear \textbf{M}odulation (F-FiLM), which uses encoded force histories to modulate visual--language representations through learned affine transformations. This mechanism retains the semantic priors of the pretrained backbone while promoting the use of force information for trajectory and contact-force prediction. Second, inspired by FoAR~\cite{2024FoAR}, we develop a contact-gated fast--slow control architecture. The policy jointly predicts nominal action and desired-force chunks, and separately estimates a contact probability. A slow position-control branch tracks the nominal motion trajectory, while a fast admittance-control branch regulates physical interaction at a higher frequency. Trajectory densification and smooth contact-gated fusion coordinate the two branches and reduce discontinuities during transitions between non-contact and contact phases. Together, these designs combine force-aware trajectory generation with responsive physical regulation for contact-rich manipulation.
In summary, our main contributions are as follows:

\begin{itemize}
    \item We propose TAO-Force, a force-conditioned VLA adaptation framework that integrates physical-interaction feedback into policy learning and execution while preserving a frozen pretrained visual--language backbone.

   \item We introduce F-FiLM to condition pretrained visual--language representations on force feedback, together with contact-causal force supervision that masks force-chunk targets at pre-contact anchors, preventing future contact forces from supervising pre-contact predictions.

    \item We develop a contact-gated fast--slow control architecture that uses the predicted contact probability to switch between nominal position control during non-contact phases and high-frequency admittance control during
    contact, with reference processing and command fusion ensuring responsive regulation and smooth recovery to nominal motion.

    \item Offline analyses and real-world evaluations on five manipulation tasks demonstrate effective force-aware adaptation and robust generalization. Across the four contact-rich execution tasks, TAO-Force achieves average success rates of 80.0\% and 64.2\% under in-domain (ID) and out-of-domain (OOD) conditions, respectively, outperforming the evaluated general-purpose VLA baselines on average.
\end{itemize}

\section{Related Work}
\subsection{General-Purpose VLA Models}
Early robotic systems typically decoupled large language or vision--language models from low-level policies, limiting their generalization beyond predefined skills and demonstrations. End-to-end VLAs subsequently unified vision, language, and action: CLIPort~\cite{shridhar2021cliport} pioneered vision--language-conditioned manipulation, while Gato~\cite{reed2022a} and VIMA~\cite{jiang2023vima} modeled multimodal inputs and actions as token sequences using Transformers. RT-1~\cite{brohan2022rt} scaled this paradigm to diverse real-world tasks, and RT-2~\cite{zitkovich2023rt} established the use of Internet-pretrained VLM backbones. RT-X~\cite{open_x_embodiment_rt_x_2023} extended training across robot embodiments, while OpenVLA~\cite{kim2024openvla} provided an open-source VLM-based policy. More recently, Octo~\cite{octo_2023}, RDT-1B~\cite{liu2024rdt}, and \ensuremath{\pi_0}~\cite{black2024pi_0} adopted diffusion or flow matching for continuous action generation. Hierarchical systems such as RT-H~\cite{rth2024arxiv}, \ensuremath{\pi_{0.5}}~\cite{intelligence2025pi05}, and GR00T N1~\cite{bjorck2025gr00t} further bridge high-level semantic reasoning and fine-grained motor execution through intermediate action representations and continuous controllers.

\subsection{Contact-Rich Manipulation Models}

Force and tactile sensing provide direct observations of physical interaction and have been increasingly incorporated into contact-rich manipulation models. At the perception and policy-learning level, early visuo-tactile methods fuse vision and touch through cross-modal attention or spatially aligned representations, as demonstrated by VTT and 3D-ViTac \cite{pmlr-v205-chen23d,huang3d}. Recent force-aware policies, including TA-VLA, ForceVLA, FoAR, and ManipForce, integrate joint torque or end-effector wrench with visual and proprioceptive features through token-based fusion, mixture-of-experts architectures, contact-aware gating, or frequency-aware encoding \cite{zhang2025elucidating,yu2026forcevla,2024FoAR,lee2026manipforce}. FD-VLA instead distills force-supervised representations from visual observations and robot states, removing the need for physical force sensing at inference time \cite{fdvla2026}. Tactile-language and tactile-VLA models further connect localized contact observations with semantic representations for contact-sensitive reasoning and action generation \cite{tla2026,Huang2025TactileVLA}. A recurring difficulty in this line of work is that force and tactile signals are acquired at rates far higher than vision, so policies must learn from asynchronous, rate-mismatched streams \cite{lee2026manipforce}.

At the execution level, force-aware perception alone may be insufficient when low-frequency position predictions are directly tracked without explicit force regulation. ForceMimic, ForceVLA2, and Tactile-VLA jointly predict pose and wrench or force targets for hybrid force--position control \cite{liu2025forcemimic,li2026forcevla2,Huang2025TactileVLA}, while CompliantVLA-adaptor regulates interaction through variable impedance control \cite{zhang2026compliantvla}. Reactive Diffusion Policy, AT-VLA, and PhaForce instead decouple slow trajectory generation from a fast tactile- or force-conditioned stream that refines actions close to control rate \cite{xue2025rdp,li2026atvla,wang2026phaforce}. Despite this progress, balancing physical-feedback conditioning with the preservation of pretrained semantic representations, while translating low-rate policy predictions into stable high-rate contact regulation, remains an important design challenge.

\section{Preliminaries and Problem Formulation}

We present the admittance-control interface used for compliant execution and formulate force-aware adaptation of a pretrained VLA policy.

\subsection{Admittance Control}

Admittance control enables compliant interaction on position-controlled robots by mapping force-tracking errors to Cartesian motion corrections. Let \ensuremath{\mathbf{x}} and \ensuremath{\mathbf{x}_d} denote the actual and reference end-effector poses, respectively, and let \ensuremath{\mathbf{F}^{\mathrm{ext}}} and \ensuremath{\mathbf{F}^{d}} denote the measured external force and desired force. The measured external force is compensated for the gravitational wrench of the end-effector tool and expressed in the controller frame. We adopt the standard second-order admittance model:
\begin{equation}
\label{eq:admittance}
\mathbf{M}(\ddot{\mathbf{x}}-\ddot{\mathbf{x}}_d)
+
\mathbf{D}(\dot{\mathbf{x}}-\dot{\mathbf{x}}_d)
+
\mathbf{K}(\mathbf{x}-\mathbf{x}_d)
=
\mathbf{F}^{\mathrm{ext}}-\mathbf{F}^{d},
\end{equation}
where \ensuremath{\mathbf{M}}, \ensuremath{\mathbf{D}}, and \ensuremath{\mathbf{K}} are the virtual mass, damping, and stiffness matrices, respectively. This model converts the force error into compliant end-effector motion for closed-loop contact regulation.

\subsection{Problem Formulation}

We consider adapting a pretrained VLA policy to contact-rich manipulation,  where successful execution requires both awareness of physical interaction and responsive contact regulation. A standard VLA policy maps a visual observation \ensuremath{o_t}, a language instruction \ensuremath{l}, and a robot state \ensuremath{\mathbf{s}_t} to an action chunk:
$$
\label{eq:vla_policy}
\pi_{\theta}(o_t,l,\mathbf{s}_t)
\mapsto
\mathbf{a}_{t:t+K},
$$
where \ensuremath{K} denotes the prediction horizon.

This formulation presents two limitations in contact-rich manipulation. The \textit{perception gap} refers to the absence of direct observations of robot--environment interaction in the input modalities of a standard VLA policy. The \textit{control gap} refers to the mismatch between low-frequency VLA action prediction and the fast contact dynamics that require real-time regulation.

To bridge these gaps, we augment the pretrained VLA policy with external-force context $\mathcal{F}^{\mathrm{ext}}_t$ as an additional input:
\begin{equation}
\label{eq:force_adaptation}
\pi_{\theta,\phi}
\left(
o_t,l,\mathbf{s}_t,\mathcal{F}^{\mathrm{ext}}_t
\right)
\mapsto
\left(
\mathbf{a}_{t:t+K},
\mathbf{F}^{d}_{t:t+K},
p^c_t
\right),
\end{equation}
The outputs \ensuremath{\mathbf{a}_{t:t+K}}, \ensuremath{\mathbf{F}^{d}_{t:t+K}}, and \ensuremath{p^c_t} denote the predicted action chunk, desired-force chunk, and contact probability, respectively. The parameters \ensuremath{\theta} are inherited from the pretrained VLA policy, while \ensuremath{\phi} denotes the parameters introduced for force-aware adaptation.

\begin{figure*}[htbp]
  \centering
  \includegraphics[width=\textwidth]{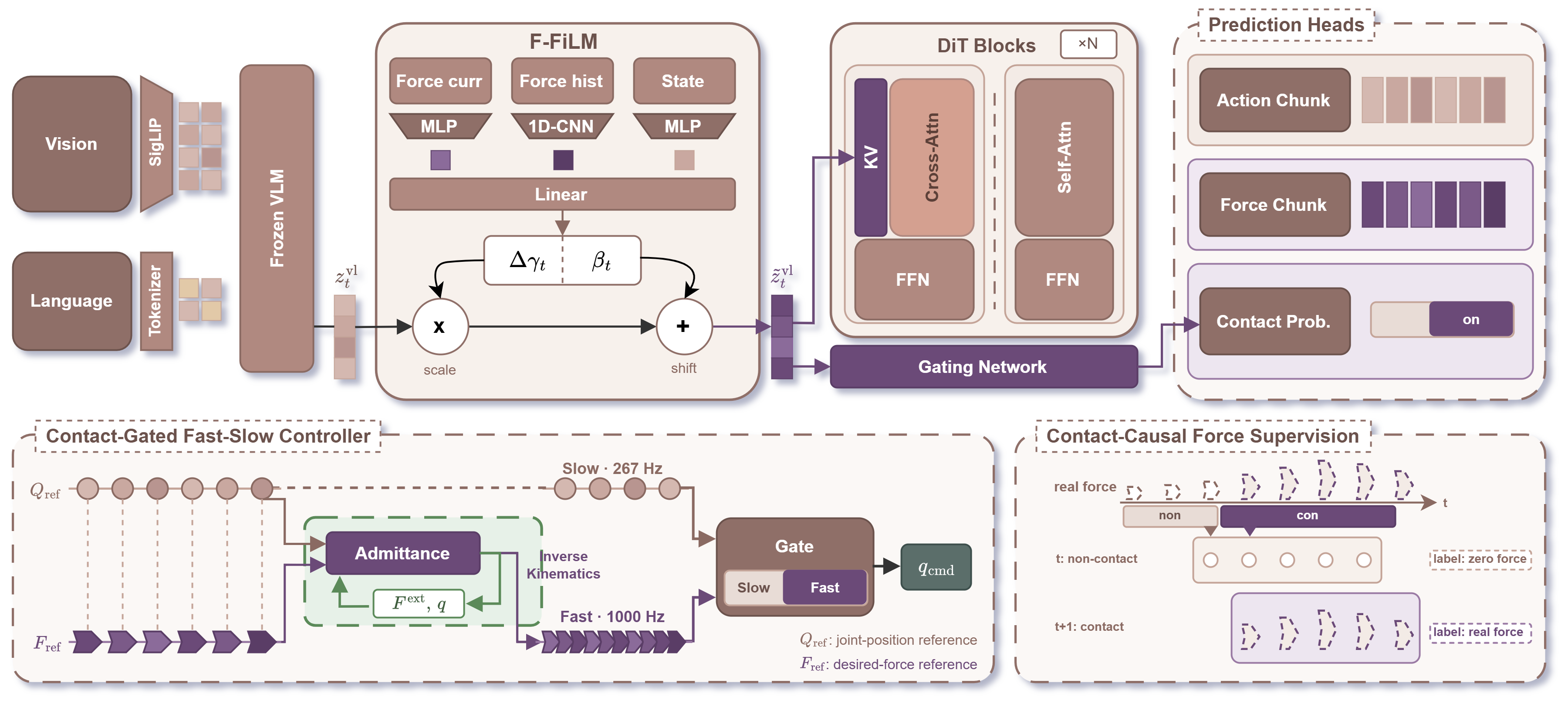}
  \caption{\textbf{Overview of TAO-Force.} TAO-Force preserves a frozen pretrained visual--language backbone and uses F-FiLM to condition its representations on force feedback, enabling the policy to jointly generate nominal action and desired-force chunks while separately estimating the current contact probability. During execution, the predicted contact probability gates between position-controlled motion
  in the non-contact phase and compliant force regulation in the contact phase. The contact-gated fast--slow controller coordinates low-frequency policy inference with high-frequency admittance control, enabling smooth transitions and responsive physical regulation for contact-rich manipulation.}
  \label{fig:architecture}
\end{figure*}

\section{The TAO-Force Framework}

As shown in Fig.~\ref{fig:architecture}, TAO-Force consists of a force-conditioned policy that jointly predicts nominal actions and desired forces while separately estimating a contact probability, together with a contact-gated fast--slow controller for responsive execution.


\subsection{Pretrained VLA Architecture}
TAO-Force builds on GR00T N1.5~\cite{nvidia2025gr00tn15}, whose Eagle visual--language backbone combines a SigLIP visual encoder with a Qwen-based language model to produce multimodal representations for a diffusion transformer (DiT) action head. We freeze the pretrained backbone to preserve its semantic and action-generation priors, and introduce lightweight force encoders, F-FiLM modulation, and additional prediction heads for force-aware adaptation.

\subsection{Force-conditioned Policy Learning}
\label{sec:force_conditioned_policy_learning}

To bridge the \textit{perception gap}, TAO-Force introduces three modifications to the pretrained VLA architecture. First, force feedback is integrated into the visual-language features as an explicit condition for DiT denoising. Second, a gating network predicts the contact probability from the force-conditioned features. Third, the DiT jointly predicts an action chunk and a desired-force chunk, providing motion and force references for the contact-gated fast--slow controller.

\paragraph{Force-conditioned modulation}
TAO-Force first extracts a visual-language representation from the pretrained VLA backbone:
$$
\mathbf{z}^{\mathrm{vl}}_t
=
f^{\mathrm{vl}}_{\theta}
\left(
o_t,l
\right).
$$
Here, the visual-language encoder is inherited from the pretrained VLA backbone. During training, the backbone parameters are kept frozen, while all newly introduced force encoders, modulation networks, and prediction heads, collectively parameterized by $\phi$, are optimized.

The current force, recent force history, and robot state are separately encoded and concatenated as the conditioning input to F-FiLM:
$$
\mathbf{e}_t
=
\operatorname{Concat}
\left[
g^{\mathrm{cur}}_{\phi}\left(\mathbf{F}^{\mathrm{ext}}_t\right),
g^{\mathrm{hist}}_{\phi}\left(\mathbf{F}^{\mathrm{ext}}_{t-H:t}\right),
g^{\mathrm{state}}_{\phi}\left(\mathbf{s}_t\right)
\right].
$$
Here, each encoded feature forms a conditioning token. The current-force and state encoders, \ensuremath{g^{\mathrm{cur}}_{\phi}} and \ensuremath{g^{\mathrm{state}}_{\phi}}, are implemented as MLPs, while \ensuremath{g^{\mathrm{hist}}_{\phi}} uses a lightweight 1D-CNN to capture recent contact dynamics.

To directly inject force feedback into the pretrained visual-language representation, we introduce F-FiLM, built upon FiLM~\cite{perez2018film}. Given the concatenated conditioning tokens $\mathbf{e}_t$, the modulation network $h^{\mathrm{film}}_{\phi}$ predicts a residual scale vector and a shift vector:
$$
\Delta\boldsymbol{\gamma}_t,\boldsymbol{\beta}_t
=
h^{\mathrm{film}}_{\phi}
\left(\mathbf{e}_t\right).
$$
These vectors modulate the pretrained visual-language representation through a feature-wise affine transformation:
$$
\tilde{\mathbf{z}}^{\mathrm{vl}}_t
=
\left(\mathbf{1}+\Delta\boldsymbol{\gamma}_t\right)
\odot\mathbf{z}^{\mathrm{vl}}_t
+
\boldsymbol{\beta}_t,
$$
where $\odot$ denotes element-wise multiplication and the FiLM scale is $\boldsymbol{\gamma}_t=\mathbf{1}+\Delta\boldsymbol{\gamma}_t$. By zero-initializing the final layer of the modulation network, we obtain $\Delta\boldsymbol{\gamma}_t=\mathbf{0}$ and $\boldsymbol{\beta}_t=\mathbf{0}$ at initialization. Thus, F-FiLM initially preserves the pretrained representation and progressively learns to incorporate physical interaction feedback.

\paragraph{Prediction heads}
The force-conditioned representation supports two policy outputs: parallel prediction of action and desired-force chunks, and instantaneous contact prediction. For chunk generation, we employ a shared DiT decoder with separate noised action and force tokens, preserving modality-specific generation paths while enabling bidirectional interaction between the two token sequences through shared self-attention.

Let \ensuremath{\mathbf{x}^{\mathrm{a}}_{t:t+K}(\tau)} and \ensuremath{\mathbf{x}^{\mathrm{f}}_{t:t+K}(\tau)} denote the noised action and force chunks at flow time \ensuremath{\tau}, respectively. Conditioned on the force-modulated visual-language representation and the current robot state, the shared DiT decoder produces temporally aligned features for the two branches:
$$
\mathbf{u}^{\mathrm{a}}_{t:t+K},
\mathbf{u}^{\mathrm{f}}_{t:t+K}
=
d^{\mathrm{dit}}_{\phi}
\left(
\mathbf{x}^{\mathrm{a}}_{t:t+K}(\tau),
\mathbf{x}^{\mathrm{f}}_{t:t+K}(\tau)
\;\middle|\;
\tilde{\mathbf{z}}^{\mathrm{vl}}_t,
\mathbf{s}_t,
\tau
\right).
$$

Two lightweight output heads predict the corresponding velocity fields:
$$
\hat{\mathbf{v}}^{\mathrm{a}}_{t:t+K}
=
q^{\mathrm{a}}_{\phi}
\left(
\mathbf{u}^{\mathrm{a}}_{t:t+K}
\right),
\qquad
\hat{\mathbf{v}}^{\mathrm{f}}_{t:t+K}
=
q^{\mathrm{f}}_{\phi}
\left(
\mathbf{u}^{\mathrm{f}}_{t:t+K}
\right).
$$
At inference time, the nominal action chunk \ensuremath{\hat{\mathbf{a}}_{t:t+K}} and desired-force chunk \ensuremath{\hat{\mathbf{F}}^{d}_{t:t+K}} are recovered by integrating the learned velocity fields from initial noise over \ensuremath{\tau \in [0,1]}. 

Contact prediction is instead formulated as an instantaneous gating problem. A lightweight classification head directly predicts the contact probability from the force-conditioned representation:
$$
\hat{p}^{c}_t
=
d^{\mathrm{c}}_{\phi}
\left(
\tilde{\mathbf{z}}^{\mathrm{vl}}_t
\right).
$$
This allows the contact head to respond directly to the current force-conditioned perceptual state, while the DiT decoder focuses on temporally coherent action and desired-force generation.

The resulting nominal action chunk \ensuremath{\hat{\mathbf{a}}_{t:t+K}}, desired-force chunk \ensuremath{\hat{\mathbf{F}}^{d}_{t:t+K}}, and contact probability \ensuremath{\hat{p}^{c}_t} constitute the complete policy-level output of TAO-Force.

\subsection{Training Objective}
\label{sec:training_objective}

We train TAO-Force on supervised demonstrations using a composite objective that combines action and force flow matching with contact classification. To ground force prediction in physical interaction and discourage reliance on non-force shortcuts, we further introduce contact-causal force supervision and force-attentive modality dropout.

\paragraph{Contact-causal force supervision}
The force supervision is derived from the six-dimensional force/torque readings recorded during demonstrations, which capture the interaction force at the robot end effector. A straightforward approach would be to directly use future segments of these measurements as force-chunk targets. However, a nonzero desired-force command is physically meaningful only after contact has been established. If a future force segment containing contact forces is used as the target while the robot is still out of contact, the model may learn to anticipate force from visual cues or demonstration-specific temporal patterns rather than grounding its prediction in current contact evidence. During deployment, even slight variations in contact onset can invalidate these learned correlations, causing the policy to predict nonzero forces before contact or to delay the required force prediction after contact occurs.

To address this issue, we gate force supervision by the current contact state. Let $c_t \in \{0,1\}$ denote whether the robot is in a task-relevant contact interval at time $t$, and let $\mathbf{F}^{\star}_{t:t+K}$ be the raw force chunk from the demonstration. The contact-causal force target is defined as
\begin{equation}
\bar{\mathbf{F}}^{\star}_{t:t+K}
=
c_t \mathbf{F}^{\star}_{t:t+K}.
\end{equation}
The force loss is evaluated over all frames, using zero targets to suppress force prediction before contact and the recorded force chunks to supervise the desired force profile during contact.

\paragraph{Force-attentive training}
A practical challenge in force-conditioned learning is that the policy may exploit visual and proprioceptive shortcuts instead of using force feedback. This is especially undesirable in contact-rich manipulation, where visually similar observations can correspond to different physical interaction states. To mitigate this issue, we apply force-attentive modality dropout during training. Specifically, we annotate force-critical temporal segments in the demonstrations and partially mask visual and proprioceptive features within these segments. This weakens the model's reliance on non-force modalities and encourages it to use force-conditioned cues for downstream prediction.

\paragraph{Overall objective}
The overall objective jointly optimizes action flow matching, force flow matching, and contact prediction. The action flow-matching loss preserves the nominal action-generation capability of the pretrained VLA policy by matching the predicted velocity field to the target velocity field induced by the ground-truth action chunk:
\begin{equation}
\mathcal{L}_{\mathrm{act}}
=
\mathbb{E}_{t,\tau}
\left[
\left\|
\hat{\mathbf{v}}^{\mathrm{a}}_{t:t+K}(\tau)
-
\mathbf{v}^{\mathrm{a},\star}_{t:t+K}(\tau)
\right\|_2^2
\right].
\end{equation}
Similarly, the force flow-matching loss matches the predicted force velocity field to the target velocity field induced by the contact-causal force target:
\begin{equation}
\mathcal{L}_{\mathrm{force}}
=
\mathbb{E}_{t,\tau}
\left[
\left\|
\hat{\mathbf{v}}^{\mathrm{f}}_{t:t+K}(\tau)
-
\bar{\mathbf{v}}^{\mathrm{f},\star}_{t:t+K}(\tau)
\right\|_2^2
\right].
\end{equation}

Given the predicted contact probability, the contact classification loss is defined using binary cross-entropy:
\begin{equation}
\mathcal{L}_{\mathrm{contact}}
=
-
\mathbb{E}_{t}
\left[
c_t
\log
\hat{p}^{c}_t
+
\left(
1-c_t
\right)
\log
\left(
1-\hat{p}^{c}_t
\right)
\right].
\end{equation}
The complete objective is
\begin{equation}
\mathcal{L}
=
\mathcal{L}_{\mathrm{act}}
+
\lambda_{\mathrm{force}}
\mathcal{L}_{\mathrm{force}}
+
\lambda_{\mathrm{contact}}
\mathcal{L}_{\mathrm{contact}},
\end{equation}
where $\lambda_{\mathrm{force}}$ and $\lambda_{\mathrm{contact}}$ balance force prediction and contact prediction relative to action generation, respectively.

\subsection{Contact-Gated Fast--Slow Control}
\label{sec:contact_gated_fast_slow_control}
\begin{algorithm}[t]
\caption{Contact-Gated Fast--Slow Control}
\label{alg:fast_slow_control}
\begin{algorithmic}[1]
\Require Predicted action chunk $\hat{\mathbf{a}}_{t:t+K}$,
force chunk $\hat{\mathbf{F}}^d_{t:t+K}$,
contact probability $\hat{p}^c_t$, and threshold $\eta$
\Ensure Joint-position commands $\mathbf{q}^{\mathrm{cmd}}$

\State $\mathbf{Q}^{\mathrm{ref}}
\gets \Call{ActionProcessor}{\hat{\mathbf{a}}_{t:t+K}}$

\State $\mathbf{F}^{\mathrm{ref}}
\gets \Call{ForceProcessor}{\hat{\mathbf{F}}^d_{t:t+K}}$

\State $g\gets\mathbb{I}[\hat{p}^c_t\geq\eta]$

\For{each aligned reference
$(\mathbf{q}^{\mathrm{pos}},\mathbf{F}^{d})$}

    \If{$g=1$} \Comment{High-freq. force control}
        \State Read $\mathbf{q}$ and $\mathbf{F}^{\mathrm{ext}}$
        \State $\dot{\mathbf{x}}^{\mathrm{adm}}
        \gets
        \Call{Admittance}
        {\mathbf{q}^{\mathrm{pos}},
         \mathbf{F}^{d},
         \mathbf{F}^{\mathrm{ext}},
         \mathbf{q}}$
        \State $\dot{\mathbf{q}}^{\mathrm{adm}}
        \gets
        \Call{IK}
        {\dot{\mathbf{x}}^{\mathrm{adm}},\mathbf{q}}$
        \State $\mathbf{q}^{\mathrm{cmd}}
        \gets
        \mathbf{q}^{\mathrm{cmd},-}
        +\Delta t\,\dot{\mathbf{q}}^{\mathrm{adm}}$
        \State $b_f\gets\mathrm{false}$

    \Else \Comment{Low-freq. Position control}
        \If{$g^{-}=1$}
            \State $\Delta\mathbf{q}
            \gets
            \mathbf{q}^{\mathrm{cmd},-}-\mathbf{q}^{\mathrm{pos}}$
            \State $s\gets0$, $\quad b_f\gets\mathrm{true}$
        \EndIf

        \If{$b_f=\mathrm{true}$}
            \State $w(s)\gets1-10s^3+15s^4-6s^5$
            \State $\mathbf{q}^{\mathrm{cmd}}
            \gets
            \mathbf{q}^{\mathrm{pos}}+w(s)\Delta\mathbf{q}$
            \State Update $s$ toward $1$
        \Else
            \State $\mathbf{q}^{\mathrm{cmd}}
            \gets\mathbf{q}^{\mathrm{pos}}$
        \EndIf
    \EndIf

    \State Send $\mathbf{q}^{\mathrm{cmd}}$ to the robot
    \State $\mathbf{q}^{\mathrm{cmd},-}\gets\mathbf{q}^{\mathrm{cmd}}$,
    $\quad g^{-}\gets g$
\EndFor

\end{algorithmic}
\end{algorithm}
To bridge the \textit{control gap} between chunk-level policy predictions
and high-frequency robot commands, TAO-Force employs the contact-gated
fast--slow controller summarized in
Algorithm~\ref{alg:fast_slow_control}. The predicted contact probability
is thresholded to select between nominal position control in free space
and high-frequency admittance control during contact.

\paragraph{Reference processing}
Following VLA-RAIL~\cite{zhao2025vlarailrealtimeasynchronousinference},
the action processor first densifies each predicted action chunk using
polynomial fitting and then applies a minimum-jerk transition between
successive chunks. This produces a smooth joint-reference sequence
\ensuremath{\mathbf{Q}^{\mathrm{ref}}}. At each chunk boundary, the force processor maps the action target index to the force timeline and smooths the beginning of the new force chunk against the currently executed force reference. The blending weight progressively shifts from the current reference to the new prediction, reducing discontinuities
and producing \ensuremath{\mathbf{F}^{\mathrm{ref}}}.

\paragraph{Contact-gated control}
During contact, the admittance controller uses the nominal joint
reference as a weak positional anchor while regulating the measured
external force toward the desired force. A small stiffness limits
accumulated positional drift without dominating force regulation. The
resulting Cartesian velocity is mapped to joint space and integrated at
the high-frequency control rate, as detailed in
Algorithm~\ref{alg:fast_slow_control}.

When contact ends, the force-controlled command may deviate from the
nominal trajectory. Directly switching back to position control could
therefore cause an abrupt motion. TAO-Force gradually removes this
command residual using a quintic smoothstep weight, smoothly reconnecting
the robot to the nominal trajectory. If contact recurs during this
transition, trajectory fusion is canceled and force control immediately
resumes.

\begin{figure}[t]
    \centering
    \includegraphics[width=\linewidth]{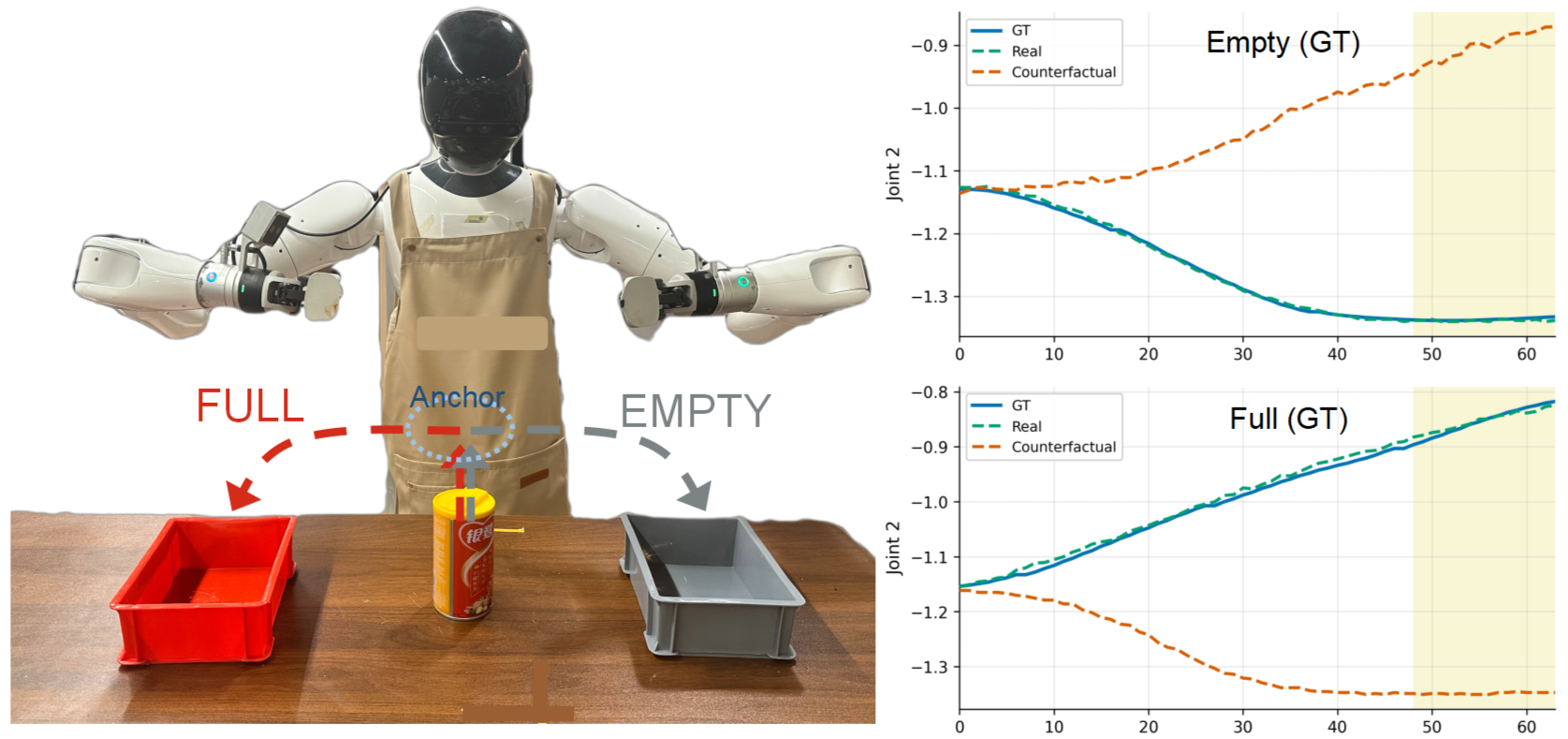} 
    \caption{\textbf{Effectiveness of Force Perception.} \textit{Left:} The Blind Box Sorting setup, highlighting the decision anchor at which the trajectories diverge based on weight. \textit{Right:} The generated action chunks (Joint 2) at the anchor under real and counterfactual force conditions, with the yellow shaded area indicating the final 16 frames of each chunk used for evaluation.}
    
    \label{fig:counterfactual}
\end{figure}

\section{Experiments}
\label{sec:experiments}

The experimental evaluation is designed to comprehensively assess TAO-Force's ability to bridge the \textit{perception gap} and \textit{control gap} in contact-rich manipulation. Through both offline analyses and real-world robot deployments, we aim to answer the following core research questions:
\begin{itemize}
    \item \textbf{Q1:} Does force feedback meaningfully influence trajectory generation?
    \item \textbf{Q2:} Does TAO-Force outperform VLA baselines in task success and robustness under both ID and OOD conditions?
    \item \textbf{Q3:} How does each key design contribute to task success?
\end{itemize}

\subsection{Experimental Setups}
\label{sec:experimental_setup}

\paragraph{Hardware Platform} All real-world experiments are conducted on the AgiBot G1 dual-arm humanoid robot~\cite{ZhiyuanG1}. Visual observations are captured by three built-in RGB cameras (one head-mounted and two wrist-mounted), while force feedback is acquired directly from the six-axis force/torque sensors at the wrists.

The system is deployed with a hierarchical fast--slow control architecture. The VLA policy runs at 5 Hz on a single NVIDIA RTX 5070 Laptop GPU with 8 GB VRAM, with observations buffered at 30 Hz. The predicted 30 Hz action and force chunks are processed into 267 Hz position and force references, respectively, while the admittance controller then incorporates real-time force and joint-position measurements to generate compliant commands at 1000 Hz.

\paragraph{Experimental Tasks}
We evaluate force perception, contact-rich execution, and OOD robustness on five real-world manipulation tasks, with 30 trials per task in each of the ID and OOD settings.

In \textbf{Blind Box Sorting}, the robot sorts visually identical canned porridge by weight, and succeeds if each can is placed in the correct bin. In \textbf{Peeling Cucumber}, the robot peels along the cucumber surface, with success defined as removing a continuous peel longer than half the cucumber's length. In \textbf{Pressing Pump Head}, the robot presses a soap-pump head, and succeeds if liquid is dispensed without excessive force. In \textbf{Wiping Whiteboard Marks}, the robot uses an eraser to wipe target marks, with success determined by their visible removal. In \textbf{Cutting Cucumber}, the robot uses a knife to cut a cucumber, and succeeds if it cuts completely through the cucumber.

\paragraph{Baselines}
\label{sec:baselines}
We compare TAO-Force against two state-of-the-art VLA baselines, GR00T N1.7 and \textbf{\ensuremath{\pi_{0.5}}}. We exclude Blind Box Sorting from the baseline evaluation because distinguishing visually identical objects by weight inherently requires force perception, making baselines without force input inapplicable to this task.

\begin{figure}[t]
    \centering
    \includegraphics[width=\linewidth]{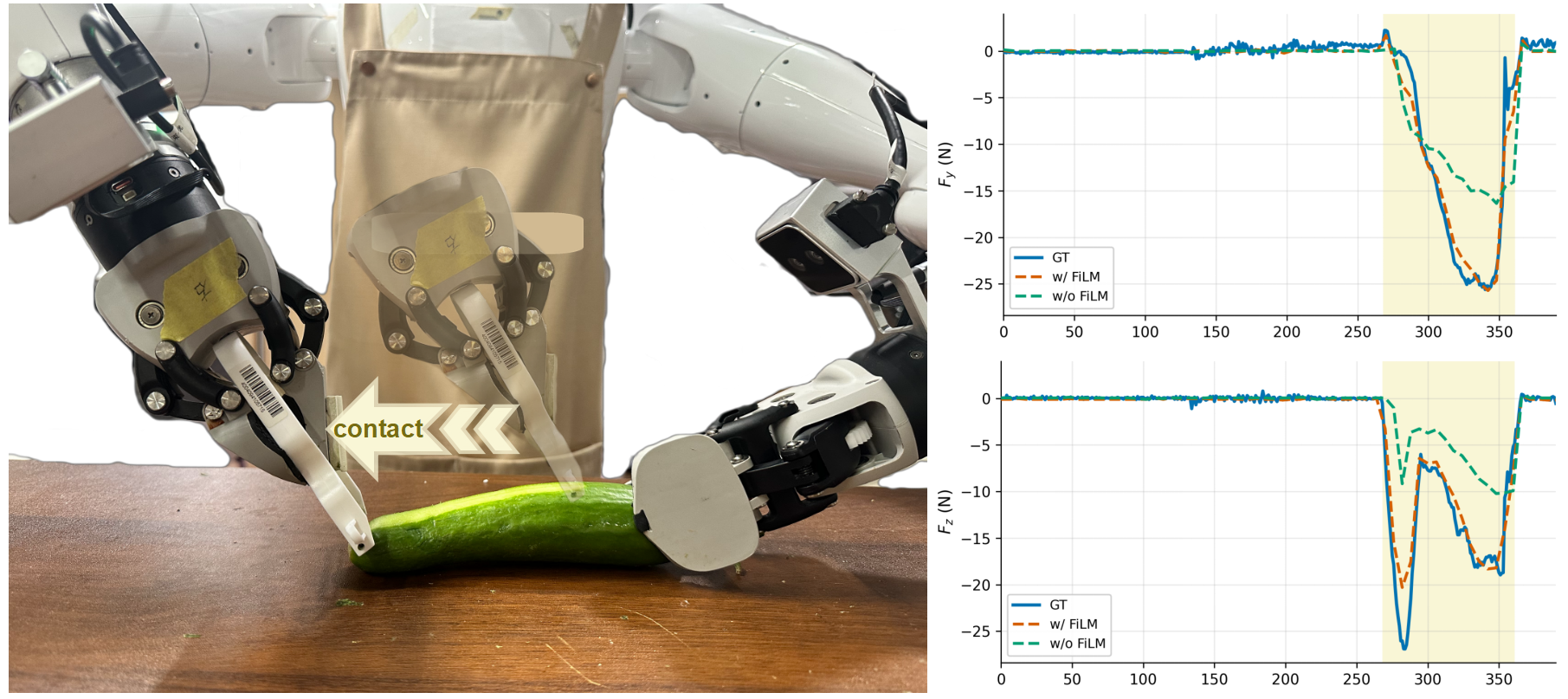}
    \caption{\textbf{Contact-Force Prediction Accuracy.}
    \textit{Left:} Illustration of the contact phase.
    \textit{Right:} Predicted and ground-truth force trajectories, with yellow regions indicating contact-dominant intervals inferred from the predicted contact probability.
    \ensuremath{F_y} and \ensuremath{F_z} are shown as they exhibit the largest variations.}
    \label{fig:force_fitting}
\end{figure}


\begin{table}[htbp]
\centering
\caption{Results of counterfactual force injection.}
\label{tab:counterfactual_quant}
\resizebox{\linewidth}{!}{
\begin{tabular}{lllcc}
\toprule
\textbf{Episode}
& \textbf{Method}
& \textbf{Force Input}
& \textbf{Trajectory Error}
& \textbf{Flip Rate} \\
\midrule

\multirow{6}{*}{Empty}
& \multirow{2}{*}{F-FiLM (Ours)}
& Empty & 0.0001 & -- \\
& & Full
& 0.0186\rlap{\quad\textcolor{green!70!black}{
\scriptsize 0.0185\ensuremath{\uparrow}}}
& \textbf{90\%} \\
\cmidrule(lr){2-5}

& \multirow{2}{*}{DePost}
& Empty & 0.0005 & -- \\
& & Full & 0.0056 & 20\% \\
\cmidrule(lr){2-5}

& \multirow{2}{*}{FVLMoE}
& Empty & 0.0005 & -- \\
& & Full & 0.0005 & 0\% \\
\midrule

\multirow{6}{*}{Full}
& \multirow{2}{*}{F-FiLM (Ours)}
& Full & 0.0001 & -- \\
& & Empty
& 0.0302\rlap{\quad\textcolor{green!70!black}{
\scriptsize 0.0301\ensuremath{\uparrow}}}
& \textbf{100\%} \\
\cmidrule(lr){2-5}

& \multirow{2}{*}{DePost}
& Full & 0.0002 & -- \\
& & Empty & 0.0326 & 90\% \\
\cmidrule(lr){2-5}

& \multirow{2}{*}{FVLMoE}
& Full & 0.0001 & -- \\
& & Empty & 0.0036 & 10\% \\
\midrule

\multicolumn{5}{l}{
\scriptsize \textit{Note:} DePost and FVLMoE are adapted from
TA-VLA~\cite{zhang2025ta} and ForceVLA~\cite{yu2026forcevla}, respectively.} \\
\bottomrule
\end{tabular}
}
\end{table}

\begin{table*}[htbp]
\centering
\caption{REAL-WORLD EVALUATION SUCCESS RATES.}
\label{tab:real_world_success}
\resizebox{\linewidth}{!}{
\begin{tabular}{
lcc:cccccccc
>{\columncolor{gray!15}}c
>{\columncolor{gray!15}}c
}
\toprule
\multirow{3}{*}{\textbf{Model}}
& \multicolumn{2}{c:}{\textbf{Blind Box Sorting}}
& \multicolumn{2}{c}{\textbf{Peeling Cucumber}}
& \multicolumn{2}{c}{\textbf{Pressing Pump Head}}
& \multicolumn{2}{c}{\textbf{Wiping Whiteboard Marks}}
& \multicolumn{2}{c}{\textbf{Cutting Cucumber}}
& \multicolumn{2}{>{\columncolor{gray!15}}c}{\textbf{Average}\ensuremath{^\dagger}} \\
\cmidrule(lr){2-3}
\cmidrule(lr){4-5}
\cmidrule(lr){6-7}
\cmidrule(lr){8-9}
\cmidrule(lr){10-11}
\cmidrule(lr){12-13}
& \textbf{ID} & \textbf{OOD}
& \textbf{ID} & \textbf{OOD}
& \textbf{ID} & \textbf{OOD}
& \textbf{ID} & \textbf{OOD}
& \textbf{ID} & \textbf{OOD}
& \textbf{ID} & \textbf{OOD} \\
& & Flickering Light
& & Table Height -3 cm
& & Bottle Height +2.5 cm
& & Different Eraser
& & Different Knife
& & \\
\midrule
GR00T N1.7
& N/A & N/A
& 76.7\% & 26.7\%
& 63.3\% & 23.3\%
& 53.3\% & 30.0\%
& 60.0\% & 0.0\%
& 63.3\% & 20.0\% \\

\ensuremath{\pi_{0.5}}
& N/A & N/A
& 53.3\% & 53.3\%
& 63.3\% & 16.7\%
& \textbf{66.7\%} & 36.7\%
& 63.3\% & 10.0\%
& 61.7\% & 29.2\% \\

TAO-Force (Ours)
& \textbf{100.0\%} & \textbf{100.0\%}
& \textbf{96.7\%} & \textbf{86.7\%}
& \textbf{80.0\%} & \textbf{70.0\%}
& 60.0\% & \textbf{53.3\%}
& \textbf{83.3\%} & \textbf{46.7\%}
& \textbf{80.0\%} & \textbf{64.2\%} \\
\bottomrule
\end{tabular}
}

\vspace{2pt}
\begin{minipage}{\linewidth}
\footnotesize
\ensuremath{^\dagger} Average success rates are computed over the four tasks excluding Blind Box Sorting.
\end{minipage}
\end{table*}

\subsection{Analysis of Force-Aware Trajectory Generation}
\label{sec:counterfactual}

To answer \textbf{Q1}, we focus on Blind Box Sorting, which provides a particularly suitable testbed for assessing the effect of force perception. We therefore conduct an offline counterfactual analysis on this task to examine whether TAO-Force genuinely conditions its trajectory generation on force feedback rather than on visual cues alone.

\textbf{Evaluation Protocol.} We use 20 test trajectories, equally split between empty and full cans, and define the decision anchor as the timestep at which their routing paths begin to diverge (Fig.~\ref{fig:counterfactual}). At this anchor, we fix the visual observation and provide either the factual force profile or a counterfactual one obtained by swapping the force signatures of empty and full cans. We evaluate the final 16 frames of the generated Joint 2 action chunk, where the routing difference is most pronounced. Trajectory Error is the MSE relative to the original ground-truth trajectory, and Flip Rate is the percentage of trajectories redirected according to the injected force. We compare F-FiLM against two existing force-feature fusion methods: DePost, proposed in TA-VLA~\cite{zhang2025ta}, and FVLMoE, proposed in ForceVLA~\cite{yu2026forcevla}.

\textbf{Results and Analysis.} As shown in Fig.~\ref{fig:counterfactual}, F-FiLM closely reproduces the ground-truth trajectory under factual force input, while counterfactual force injection redirects the generated trajectory toward the routing direction associated with the injected force. Table~\ref{tab:counterfactual_quant} further shows that F-FiLM responds consistently in both swapping directions, whereas DePost exhibits an asymmetric response and FVLMoE shows limited sensitivity to the injected force. Overall, F-FiLM provides the most consistent force-aware trajectory adaptation among the evaluated fusion methods.

\subsection{Analysis of Contact-Force Prediction}
\label{sec:force_tracking}

To further assess TAO-Force's ability to model contact dynamics, we conduct an offline force-prediction analysis on the cucumber-peeling test trajectories.

\textbf{Evaluation Protocol.} We identify contact-dominant intervals by thresholding the predicted contact probability at 0.8 and uniformly sample temporal anchors within these intervals. At each anchor, we evaluate the right-wrist force prediction at the first frame of the predicted chunk. We report MSE separately for the translational force components \ensuremath{(F_x, F_y, F_z)} and rotational torque components \ensuremath{(T_x, T_y, T_z)}.

\textbf{Results and Analysis.} As illustrated in Fig.\ref{fig:force_fitting}, the predicted force and torque signals closely follow the rapid variations in the ground-truth signals during peeling contact.

As shown in Table \ref{tab:cucumber_film_contact}, F-FiLM substantially improves prediction accuracy across both force and torque components.

\begin{table}[htbp]
\centering
\caption{Contact-phase force and torque prediction errors.}
\label{tab:cucumber_film_contact}
\begin{tabular}{lcc}
\toprule
\textbf{Method} & \textbf{Force MSE} & \textbf{Torque MSE} \\
\midrule
w/o F-FiLM & 35.097 & 0.653 \\
w/ F-FiLM  & 4.098\rlap{\quad\textcolor{green!70!black}{\scriptsize 30.999\ensuremath{\downarrow}}}
           & 0.070\rlap{\quad\textcolor{green!70!black}{\scriptsize 0.583\ensuremath{\downarrow}}} \\
\bottomrule
\end{tabular}
\end{table}

\subsection{Real-World Task Performance}
\label{sec:real_world_eval}

Table~\ref{tab:real_world_success} summarizes the real-world success rates under the ID and OOD settings defined in the experimental setup, with 30 trials per setting.

For \textbf{Blind Box Sorting}, TAO-Force achieves a 100.0\% success rate under both ID and OOD conditions. As discussed in Section~\ref{sec:experimental_setup}, baselines without force input are inapplicable to this task and are therefore marked as N/A.

For \textbf{Peeling Cucumber} and \textbf{Pressing Pump Head}, the OOD perturbations both change the height of the manipulated object and thus alter the contact geometry. In Peeling Cucumber, failures mainly occur when the peeler misses the cucumber surface or removes only a short peel. In Pressing Pump Head, failures are primarily caused by excessive pressing force after the bottle height is increased. TAO-Force achieves the highest OOD success rates on both tasks.

For \textbf{Wiping Whiteboard Marks} and \textbf{Cutting Cucumber}, the OOD perturbations replace the tool used during evaluation. Tool changes alter the tool geometry, thereby shifting the contact timing and changing the resulting contact-force characteristics during interaction. This mismatch can cause insufficient wiping pressure, unstable contact, or ineffective cutting, leading to a clear success-rate drop under OOD conditions.

Overall, TAO-Force achieves the strongest OOD performance and exhibits a smaller ID-to-OOD performance drop than GR00T N1.7 and \ensuremath{\pi_{0.5}} across the evaluated tasks. These results demonstrate improved robustness to variations in contact geometry, tool properties, and environmental conditions. Together, the ID and OOD evaluations answer \textbf{Q2} affirmatively: TAO-Force consistently outperforms the VLA baselines in task success and generalizes more robustly to contact-related perturbations.

\subsection{Ablation Study}
\label{sec:ablation}

To answer \textbf{Q3}, we evaluate F-FiLM modulation and force control on two representative contact-rich tasks: Peeling Cucumber and Pressing Pump Head. GR00T N1.5 + F-FiLM incorporates force-conditioned feature modulation into the VLA backbone without force-controlled execution, whereas GR00T N1.5 + Force Control applies the proposed force-control branch without F-FiLM-based trajectory generation. TAO-Force integrates both components to combine force-conditioned trajectory generation with compliant execution.

\begin{table}[htbp]
\centering
\caption{Ablation study of real-world success rates.}
\label{tab:ablation_real_world}
\resizebox{\linewidth}{!}{
\begin{tabular}{lcccc}
\toprule
\multirow{2}{*}{\textbf{Method}}
& \multicolumn{2}{c}{\textbf{Peeling Cucumber}}
& \multicolumn{2}{c}{\textbf{Pressing Pump Head}} \\
\cmidrule(lr){2-3}
\cmidrule(lr){4-5}
& \textbf{ID} & \textbf{OOD}
& \textbf{ID} & \textbf{OOD} \\
\midrule
GR00T N1.5
& 56.7\% & 3.3\%
& 66.7\% & 26.7\% \\
GR00T N1.5 + F-FiLM
& 70.0\% & 23.3\%
& 66.7\% & 63.3\% \\
GR00T N1.5 + Force Control
& 63.3\% & 60.0\%
& 43.3\% & 33.3\% \\
TAO-Force (Ours)
& \textbf{96.7\%} & \textbf{86.7\%}
& \textbf{80.0\%} & \textbf{70.0\%} \\
\bottomrule
\end{tabular}
}
\end{table}

As shown in Table~\ref{tab:ablation_real_world}, F-FiLM and force control provide complementary benefits. F-FiLM enables measured force feedback to modulate multimodal representations during trajectory generation.

Force control instead improves execution-time adaptability through closed-loop regulation. Compared with directly tracking predicted actions, regulating the desired contact force is more tolerant to variations in contact timing and local interaction conditions. However, neither component alone performs consistently well across both tasks and evaluation settings.

By integrating F-FiLM with force control, TAO-Force achieves the highest success rate in every evaluated setting. These results answer \textbf{Q3}: force-conditioned multimodal fusion improves upstream trajectory generation, while compliant force control enhances downstream execution robustness, and their combination is necessary for the strongest overall performance.

\section{Discussion and Conclusion}
\label{sec:discussion_conclusion}

This work presents TAO-Force, a force-conditioned VLA framework for contact-rich manipulation. TAO-Force introduces F-FiLM to incorporate measured force feedback into multimodal representations, enabling force-aware trajectory generation. It further employs a fast--slow control architecture: the slow system generates action and desired-force trajectories from multimodal observations, while the fast system performs closed-loop force regulation during execution. Together, these components bridge force-aware perception and responsive control for robust contact-rich manipulation. Real-world experiments demonstrate that TAO-Force consistently outperforms the VLA baselines, while the ablation study validates the complementary contributions of F-FiLM and force control to robust contact-rich manipulation.


\bibliographystyle{ieeetr}
\bibliography{scholar}
\end{document}